%% file: deepmind.tex
\documentclass{article} % For LaTeX2e
\usepackage{iclr2020_conference,times}

\input{math_commands.tex}

\usepackage[utf8]{inputenc} % allow utf-8 input
\usepackage[T1]{fontenc}    % use 8-bit T1 fonts
\usepackage{hyperref}       % hyperlinks
\usepackage{url}            % simple URL typesetting
\usepackage{booktabs}       % professional-quality tables
\usepackage{nicefrac}       % compact symbols for 1/2, etc.
\usepackage{microtype}      % microtypography
\usepackage{pifont}

\usepackage{times}
\usepackage{enumitem}

\usepackage{bm}
\usepackage{wrapfig}
\usepackage{changepage}
\usepackage{amsmath}
\usepackage{amsthm}
\usepackage{amssymb}
\usepackage{amsfonts,eucal,amsbsy}
\usepackage{mathtools}
\usepackage{multirow}
\usepackage[normalem]{ulem}
\usepackage{natbib}

\usepackage{subfigure}
\usepackage{booktabs}
\usepackage{tablefootnote}

\newcommand{\ignore}[1]{}
\newcommand{\vect}[1]{\mathbf{#1}}
\newcommand{\set}[1]{\mathcal{#1}}
\newcommand{\struct}[1]{\boldsymbol{#1}}

\DeclareSymbolFont{extraup}{U}{zavm}{m}{n}
\DeclareMathSymbol{\varheart}{\mathalpha}{extraup}{86}
\DeclareMathSymbol{\vardiamond}{\mathalpha}{extraup}{87}

\title{A Mutual Information Maximization Perspective of Language Representation Learning}

\iclrfinalcopy

\author{Lingpeng Kong$^{\spadesuit}$, Cyprien de Masson d'Autume$^{\spadesuit}$, Wang Ling$^{\spadesuit}$, Lei Yu$^{\spadesuit}$, Zihang Dai$^{\heartsuit\clubsuit}$\\ 
\textbf{Dani Yogatama}$^{\spadesuit}$\\
DeepMind$^{\spadesuit}$, Carnegie Mellon University$^{\heartsuit}$, Google Brain$^{\clubsuit}$\\
London, United Kingdom \\
\texttt{\{lingpenk,cyprien,lingwang,leiyu,zihangd,dyogatama\}@google.com}
}

\begin{document}
\maketitle

\begin{abstract}
We show state-of-the-art word representation learning
methods maximize an objective function that is a 
lower bound on the mutual information
between different parts of a word sequence (i.e., a sentence).
%and they can be seen as instances of contrastive learning.
Our formulation provides an alternative
perspective that unifies classical word embedding models 
(e.g., Skip-gram) and modern contextual embeddings (e.g., BERT, XLNet).
In addition to enhancing 
our theoretical understanding of these methods,
our derivation leads to
a principled framework that can be used to 
construct new self-supervised tasks.
We provide an example by
drawing inspirations from related methods 
based on mutual information maximization
that have been successful in computer vision, 
and introduce a simple self-supervised objective	
that maximizes the mutual information
between a global sentence representation
and $n$-grams in the sentence.
%In addition to facilitating the creation of new self-supervised tasks,
Our analysis offers
a holistic view of representation
learning methods to
transfer knowledge and translate 
progress
across multiple domains (e.g., natural language processing, 
computer vision, audio processing).
%We believe that the mutual information maximization 
%framework provides a principled 
%foundation to design better 
%self-supervised language tasks.
\end{abstract}

%\balance
% Incude paper content from external files 

\section{Introduction}
Advances in representation learning 
have driven progress 
in natural language processing.
Performance on many downstream 
tasks have improved considerably,
achieving parity with human baselines in 
benchmark leaderboards such as SQuAD \citep{squad,squad2} and GLUE \citep{glue}.
The main ingredient is 
the ``pretrain and fine-tune'' approach,
where a large text encoder is trained 
on an unlabeled corpus
with self-supervised training objectives and used to initialize
a task-specific model.
Such an approach has also been shown to reduce
the number of training examples that is needed to achieve good
performance on the task of interest \citep{glipaper}.

In contrast to first-generation models that 
learn word type embeddings
\citep{skipgram,glove},
recent methods
have focused on contextual token representations---i.e., 
learning an encoder to 
represent words in context.
%A key difference among contextual word representation 
%methods is on the type of
%unsupervised objective functions 
%used to train their encoders 
%(in addition to the choice of encoder architectures).
Many of these encoders are trained with 
a language modeling objective, where 
the representation of a context is trained to 
be predictive of 
a target token by maximizing the log 
likelihood of predicting this token \citep{daile,ulmfit,gpt,gpt2}.
In a vanilla language modeling objective, the 
target token
is always the next token that follows the context.
\citet{elmo} propose an improvement by adding a reverse 
objective that also predicts the word token that precedes the context.
Following this trend, current state-of-the-art encoders
such as BERT \citep{bert}
and XLNet \citep{xlnet} 
%\lpkcomment{I kind of want to mention all the bert refector here ALBERT-Team Google Language	ALBERT (Ensemble) Microsoft D365 AI and UMD	Adv-RoBERTa (ensemble)Facebook AI	RoBERTa, so that we no get haters....}
are also trained with variants of the language modeling
objective: masked language modeling and permutation language modeling. %bidirectional information.

In this paper, we provide an alternative view and show 
that these methods also maximize
a lower bound on the mutual information between different
parts of a word sequence.
Such a framework is inspired by the InfoMax 
principle \citep{linsker}
and has been the main driver of
progress in self-supervised representation learning in other domains
such as computer vision, audio processing, and reinforcement learning
\citep{mine,cpc,deepinfomax,amdim,lowe}.
Many of these methods are trained to maximize a particular
lower bound called InfoNCE \citep{cpc}---also known as contrastive
learning \citep{arora}.
The main idea behind contrastive learning is to divide an input data
into multiple (possibly overlapping) views and maximize 
the mutual information
between encoded representations of these views, using
views derived from other inputs as negative samples.
In \S{\ref{sec:mim}}, we provide an overview of representation learning with
mutual information maximization. 
We then show how 
the skip-gram objective (\S{\ref{sec:skipgram}}; \citealt{skipgram}),
masked language modeling (\S{\ref{sec:bert}}; \citealt{bert}), and permutation
language modeling (\S{\ref{sec:xlnet}}; \citealt{xlnet}), fit in this framework.
%In this paper, we present a framework based
%on mutual information maximization---that unifies
%many objective functions that are used to learn text encoders.
%We show that state-of-the-art objective functions 
%(e.g., masked language modeling, next sentence
%prediction, and permutation language modeling) are instances
%of this framework. We also draw connections to
%earlier work on non-contextual word embeddings such as skip
%grams \citep{skipgram}.

%Our analysis provides an insight into how 
%state-of-the-art language
%representation learning methods are related to contrastive
%learning methods.
In addition to providing a principled theoretical 
understanding that bridges progress in multiple areas,
our proposed framework also gives rise to a 
general class of word representation learning models
%that can be instantiated with other 
%related objective functions.
which serves as a basis for designing
and combining self-supervised training objectives 
to create better language representations.
As an example, we show how to use this framework to
construct a simple self-supervised objective 
that maximizes
the mutual information between a sentence
and $n$-grams in the sentence (\S\ref{sec:infobert}).
We combine it with a variant of the masked language modeling
objective and 
show that the resulting representation performs better,
particularly
on tasks such as question answering and linguistics acceptability 
(\S\ref{sec:experiments}).
%Our approach 
%learning methods to learn contextual word representations.
%that unlock the full potential of 
%mutual information maximization
%as a language representation learning technique.

\section{Mutual Information Maximization}
\label{sec:mim}
%before discussing how Skip-gram, BERT, and XLNet fit into this framework. %(\S{\ref{sec:skipgram}}, \S{\ref{sec:bert}}, and \S{\ref{sec:xlnet}}),
%and possible extensions inspired by self-supervised
%objectives used in computer vision (\S{\ref{sec:infobert}}).

Mutual information measures dependencies between random variables.
Given two random variables $A$ and $B$, it can be understood as how much
knowing $A$ reduces the uncertainty in $B$ or vice versa.
Formally, the mutual information between $A$ and $B$ is:
\begin{align*}
I(A,B) = H(A) - H(A\mid B) = H(B) - H(B\mid A).
\end{align*}
Consider $A$ and $B$ to be different views of
an input data (e.g., a word and its context, two different partitions of a sentence). Consider a function $f$ that takes
$A=a$ and $B=b$ as its input.
The goal of training is
to learn parameters of the function $f$ that maximizes 
$I(A,B)$.

Maximizing mutual information 
directly is generally intractable when the
function $f$ consists of modern encoders 
such as neural networks \citep{paninski},
so we need to resort to a lower bound on $I(A,B)$.
One particular lower bound that has been shown to work well 
in practice is InfoNCE \citep{logeswaran,cpc},\footnote{Alternative bounds include Donsker-Vardhan representation \citep{dv} and Jensen-Shannon estimator \citep{jenshan},
but we focus on InfoNCE here.}
which is based on
Noise Contrastive Estimation (NCE; \citealp{nce}).\footnote{
See \citet{cpc,poole}
for detailed derivations of InfoNCE as a bound on mutual information.}
%Consider a discriminator $h_{\psi}: \mathcal{X} \times \mathcal{C} \rightarrow \mathbb{R}$ 
%parameterized by $\psi \in \Psi$. We have:
InfoNCE is defined as:
\begin{align}
\label{eq:nce}
I(A,B) &\geq \mathbb{E}_{p(A,B)}\left[f_{\bm{\theta}}(a,b) - \mathbb{E}_{q(\tilde{\set{B}})} \left[\log \sum_{\tilde{b} \in \mathcal{\tilde{B}}} \exp f_{\bm{\theta}}(a,\tilde{b})\right]\right] + \log \mid\mathcal{\tilde{B}}\mid,
\end{align}

% \begin{align*}
% I(f(A,B)) &\geq \mathbb{E}_{p(a,b)}\left[f_{\boldsymbol{\theta}}(a,b) - \mathbb{E}_{q(\tilde{\mathcal{B}})} \left[\log \sum_{\tilde{b} \in \mathcal{\tilde{B}}} \exp f_{\boldsymbol{\theta}}(a,\tilde{b})\right]\right] + \log \mid\mathcal{\tilde{B}}\mid,
% \end{align*}

where $a$ and $b$ are different views of an input sequence, 
$f_{\bm{\theta}} \in \mathbb{R}$ is a function parameterized by
$\bm{\theta}$ (e.g., a dot product between encoded representations of a word and its context, a dot product between encoded representations of two partitions of a sentence), and
$\mathcal{\tilde{B}}$ is a set of samples drawn 
from a proposal distribution $q(\mathcal{\tilde{B}})$.
The set $\mathcal{\tilde{B}}$ contains the positive sample $b$ and $|\mathcal{\tilde{B}}| - 1$ negative samples. 
%drawn from $q(\tilde{b})$, so $q(\tilde{\set{B}}) \propto \prod_{i=1}^{|\mathcal{\tilde{B}}| - 1} q(\tilde{b}_i)$.

Learning representations based on this objective 
is also known as contrastive learning. \citet{arora}
show representations learned by such a method have provable
performance guarantees and reduce sample complexity on downstream tasks.

We note that InfoNCE is related to cross-entropy.
%which is a commonly used loss when learning language representations.
When $\mathcal{\tilde{B}}$ always
includes all possible values of the random variable $B$ 
(i.e., $\mathcal{\tilde{B}} = \mathcal{B})$ and they are uniformly distributed, maximizing InfoNCE is 
analogous to maximizing the standard cross-entropy loss:
\begin{align}
\label{eq:crossent}
\mathbb{E}_{p(A,B)}\left[f_{\bm{\theta}}(a,b) - \log \sum_{\tilde{b} \in \mathcal{B}} \exp f_{\bm{\theta}}(a,\tilde{b})\right].
\end{align}
Eq.~\ref{eq:crossent} above shows that InfoNCE is related to
maximizing $p_{\bm{\theta}}(b\mid a)$, and it approximates
the summation over elements in $\mathcal{B}$ (i.e., the partition function)
by negative sampling.
As a function of the negative samples, the InfoNCE bound 
is tighter when $\mathcal{\tilde{B}}$ contains
more samples (as can be seen in Eq.~\ref{eq:nce} above
by inspecting the $\log |\mathcal{\tilde{B}}|$ term). 
Approximating a softmax over a large vocabulary
with negative samples is a popular technique that 
has been widely used in natural language processing in the past. 
We discuss it here to make the connection under this framework clear.

\section{Models}
\label{sec:models}
We describe how Skip-gram, BERT, and XLNet fit into the
mutual information maximization framework as instances of InfoNCE.
In the following, we assume that
%$f_{\bm{\theta}}(a,b) = g_{\bm{\omega}}(b)^{\top}g_{\bm{\psi}}(a)$ 
%or 
$f_{\bm{\theta}}(a,b) = g_{\bm{\psi}}(b)^{\top}g_{\bm{\omega}}(a)$,
where $\bm{\theta} = \{\bm{\omega},\bm{\psi}\}$.
Denote the vocabulary set by $\mathcal{V}$
and the size of the vocabulary by $V$.
For word representation learning, we seek to learn an encoder 
parameterized by $\bm{\omega}$
to represent each word in a sequence 
$\boldsymbol{x} = \{x_1, x_1, \ldots, x_T\}$ in $d$ dimensions.
%(i.e., $g_{\bm{\omega}}: V^{T} \rightarrow \mathbb{R}^{T \times d}$).
For each of the models we consider in this paper,
$a$ and $b$ are formed by taking different parts of 
$\boldsymbol{x}$ (e.g., $a \coloneqq x_0$ and $b \coloneqq x_T$).

%(\S{\ref{sec:skipgram}}, \S{\ref{sec:bert}}, and \S{\ref{sec:xlnet}}),

\subsection{Skip-gram}
\label{sec:skipgram}
%We have shown that BERT maximizes Eq.~\ref{eq:nce} 
%using different instantiations of $a$ and $b$ 
%as well as negative samples $\tilde{b}$.
%This insight allows us to show how BERT is 
%related to an earlier representation
%learning method Skip-gram \citep{word2vec}
%and a more recent one XLNet \citep{xlnet} under the
%mutual information maximization formulation.

We first start with a simple word representation 
learning model Skip-gram \citep{skipgram}.
Skip-gram is a method for learning
word representations that relies on the assumption 
that a good representation of a word 
should be predictive of its context.
The objective function that is maximized in Skip-gram 
is:
$\mathbb{E}_{p(x_i,x_j^i)}\left[p(x_j^i\mid x_i)\right]$,
where $x_i$ is a word token and $x_j^i$ is a context word of $x_i$.
%\citet{levygoldberg} show that this objective function 
%and its connection to (pointwise)
%mutual information has been studied in previous work.
%Since this is a cross-entropy objective function,
%A contrastive learning 
%variant of the skip-gram objective has been shown by
%\citet{mnihnce}.
%Here, we discuss this objective function 
%in the context of InfoNCE.

Let $b$ be the context word to be predicted $x_j^i$
and $a$ be the input word $x_i$.
Recall that $f_{\bm{\theta}}(a,b)$ is
$g_{\bm{\psi}}(b)^{\top}g_{\bm{\omega}}(a)$.
The skip-gram objective function can be written as an
instance of InfoNCE (Eq.~\ref{eq:nce}) where
$g_{\bm{\psi}}(b)$ and $g_{\bm{\omega}}(a)$
are embedding lookup functions that map each word type to $\mathbb{R}^d$.
(i.e., $g_{\bm{\psi}}(b), g_{\bm{\omega}}(a): \mathcal{V} \rightarrow \mathbb{R}^d$).

$p(x_j^i\mid x_i)$ can either be computed using a standard
softmax over the entire vocabulary or with negative 
sampling (when the vocabulary
is very large).
These two approaches correspond to different choices of $\tilde{\mathcal{B}}$. 
In the softmax approach, $\tilde{\mathcal{B}}$ is
the full vocabulary set $\mathcal{V}$
and each word in $\mathcal{V}$ is uniformly distributed.
In negative sampling, $\tilde{\mathcal{B}}$ is 
a set of negative samples drawn from e.g., 
a unigram distribution.

While Skip-gram has been widely accepted as an instance
contrastive learning \citep{skipgram,mnihnce}, we include it here
to illustrate its connection with modern approaches such as BERT and XLNet
described subsequently.
We can see that the two views of an input sentence that are considered in Skip-gram
are two words that appear in the same sentence, and they
are encoded using simple lookup functions.
%A contrastive learning 
%variant of the skip-gram objective has been shown by
%\citet{mnihnce}

%and $q(\mathcal{B})$ be a uniform distribution.
 %(i.e.,
%$g_{\bm{\psi}}(a)$ and $g_{\bm{\omega}}(b)$ are word embedding matrices
%parameterized by $\bm{\psi}$ and $\bm{\omega}$).
%Following this instantiation, the skip-gram objective function %maximizes Eq.~\ref{eq:crossent},
%which 
%is an instance of Eq.~\ref{eq:nce}.
%The above instantiation describes the Skip-gram variant
%that is trained using a standard softmax over
%the entire vocabulary, where $\tilde{\mathcal{B}}$ always contains 
%the full vocabulary set. 
%When the vocabulary
%is very large, it is sometimes necessary to
%use negative sampling \citep{skipgram}. As an instance of contrastive learning, 
%the difference between these two
%Skip-gram variants is only 
%in $\tilde{\mathcal{B}}$ (a set
%of negative samples vs. a full vocabulary set)
%and $q(\tilde{b}_i)$ (a unigram distribution vs.
%a uniform distribution).

\subsection{BERT}
\label{sec:bert}
%Skip-gram is a context independent word representation learning model
%which produces the same representation 
%for each word token of the same type regardless of the context it 
%appears in.
%Recent work has shown that contextual representations
%outperform context independent methods.
%We show that two best performing variants
%of this class of models---BERT and XLNet---can also be seen as methods
%that maximize Eq.~\ref{eq:nce}.

\citet{bert} introduce two self-supervised tasks for learning
contextual word representations: masked language modeling and next sentence prediction.
Previous work suggests that 
the next sentence prediction objective
is not necessary to train a high quality BERT encoder
and the masked language modeling appears to be the key to learn
good representations \citep{roberta,spanbert,lample2019cross}, so we focus
on masked language modeling here. However, we also show how
next sentence prediction fits into our framework in Appendix~\ref{appendix:nsp}.

%\paragraph{Masked language modeling.} 
In masked language modeling, given a sequence of word tokens of length $T$, $\boldsymbol{x} = \{x_1, \ldots, x_T\}$, BERT replaces 15\% of the tokens in the sequence with 
(i) a mask symbol 80\% of the time, 
(ii) a random word 10\% of the time, 
or (iii) its original word.
For each replaced token, it introduces a term in the
masked language modeling training objective 
to predict the original word
given the perturbed sequence $\boldsymbol{\hat{x}}_i = \{x_1, \ldots, \hat{x}_i, \ldots, x_T\}$ (i.e., the sequence
$\boldsymbol{x}$ masked at $x_i$).
This training objective can be 
written as: 
$\mathbb{E}_{p(x_i,\boldsymbol{\hat{x}}_i)}[p(x_i \mid \boldsymbol{\hat{x}}_i)]$.

Following our notation in \S{\ref{sec:mim}}, 
we have $f_{\bm{\theta}}(a,b) = g_{\bm{\psi}}(b)^{\top}g_{\bm{\omega}}(a)$.
Let $b$ be a masked word $x_i$
and $a$ be the masked sequence 
$\boldsymbol{\hat{x}}_i$.
%where $\hat{x}_i$ denotes a mask symbol.
Consider a Transformer encoder parameterized by
$\bm{\omega}$ and denote $g_{\bm{\omega}}(\boldsymbol{\hat{x}}_i) \in \mathbb{R}^d$
as a function that returns
%the output of a Transformer encoded sequence 
%is in $\mathbb{R}^{T \times d}$
%(i.e., the final hidden
%states for each token in the sequence),
the final hidden state 
corresponding to the $i$-th token 
(i.e., the masked token) after running $\boldsymbol{\hat{x}}_i$ through
the Transformer.
%such that $g_{\bm{\omega}}(a) \in \mathbb{R}^d$.
Let $g_{\bm{\psi}}: \mathcal{V} \rightarrow \mathbb{R}^d$ be 
a lookup function that 
maps each word type
into a vector and $\mathcal{\tilde{B}} = \mathcal{B}$ 
be the full vocabulary set $\mathcal{V}$.
Under this formulation, the masked language modeling objective
maximizes Eq.~\ref{eq:nce} and
different choices of masking probabilities 
can be understood as manipulating the joint distributions $p(a,b)$.
In BERT, the two views of a sentence correspond to
a masked word in the sentence and its masked context.

%In BERT, there is a 12\% chance (15\% $\times$ 80\%) that each 
%of the context token is replaced by a mask token 
%and a 1.5\% (15\% $\times$ 10\%) chance that each of the context 
%token is replaced by a random token.

%From the mutual information maximization perspective of
%optimizing Eq.~\ref{eq:nce},
%we should draw negative samples for 
%the log-sum-exp term either only for $a$ or $b$ 
%while keeping the other fixed (but not both).
%Since in masked language modeling we already draw
%use other words in the vocabulary as negative 
%samples for $\tilde{b}$, introducing some noise in 
%the context $\tilde{a}$ as well is inconsistent with 
%Eq.~\ref{eq:nce}.
%However, this could be seen as introducing a
%regularization effect that might benefit learning.

%\dycomment{probability of samping negative set and masked tokens}

\paragraph{Contextual vs. non-contextual.}
It is generally understood that the main difference between
Skip-gram and BERT is that Skip-gram learns representations
of word types (i.e., the representation for a word 
is always the same regardless of the context it appears in)
and BERT learns representations of word tokens.
%\citet{skipgram} show that Skip-gram can be optimized
%using Noise Contrastive Estimation.
We note that under our formulation for either Skip-gram or BERT, 
the encoder that we want 
to learn appears in $g_{\bm{\omega}}$, and $g_{\bm{\psi}}$
is not used after training.
We show that Skip-gram and BERT
maximizes a similar objective, and the main difference is 
in the choice of the encoder that forms $g_{\bm{\omega}}$---a context 
dependent Transformer
encoder that takes a sequence as 
its input for BERT and a simple word embedding lookup for Skip-gram.

\subsection{XLNet}
\label{sec:xlnet}

\citet{xlnet} propose a permutation
language modeling objective to learn contextual word representations.
This objective considers all possible factorization permutations 
of a joint distribution of a sentence.
Given a sentence $\boldsymbol{x} = \{x_1, \ldots, x_T\}$,
there are $T!$ ways to factorize its joint distribution.\footnote{For example, we can factorize
$p(\boldsymbol{x}) = p(x_1) p(x_2\mid x_1) \ldots, p(x_T\mid x_1, \ldots, x_{T-1}) = p(x_T) p(x_{T-1}\mid x_T) \ldots, p(x_1\mid x_T, \ldots, x_{2})$,
and many others.}
Given a sentence $\boldsymbol{x}$, denote a permutation
by $\boldsymbol{z} \in \mathcal{Z}$.
XLNet optimizes the objective function:
\vspace{-0.1cm}
\begin{align*}
\mathbb{E}_{p(\boldsymbol{x})}\left[\mathbb{E}_{p(\boldsymbol{z})}\left[\sum_{t=1}^T \log p(x_t^{\boldsymbol{z}} \mid \boldsymbol{x}_{<t}^{\boldsymbol{z}}) \right]\right].
\end{align*}
\vspace{-0.3cm}

%This permutation objective is closely related to the
%masked language modeling objective used in BERT.
As a running example,
consider a permutation order \texttt{3,1,5,2,4} for 
a sentence $x_1, x_2, x_3, x_4, x_5$.
Given the order, XLNet is only trained 
to predict the last $S$ tokens in practice.
For $S=1$, the context sequence
used for training is
$x_1, x_2 , x_3, \_, x_5$, with $x_4$ being the target word.

In addition to replacing the Transformer encoder with Transformer XL \citep{txl},
a key architectural innovation of XLNet 
is the two-stream self-attention.
In two-stream self attention, a shared encoder is used to
compute two sets of hidden representations from one original sequence.
They are called the query stream and the content stream.
In the query stream, the input sequence is masked at the target position,
whereas the content stream sees the word at the target position.
Words at future positions for the permutation order under consideration 
are also masked in both streams.
These masks are implemented as two attention mask matrices.
During training, the final hidden representation 
for a target position 
from the query stream is used to predict the target word.

Since there is only one set of encoder 
parameters for both streams, we show that 
we can arrive at
the permutation language modeling objective
from the masked language modeling objective
with an architectural change in the encoder.
Denote a hidden representation by $\mathbf{h}_t^k$,
where $t$ indexes the position 
and $k$ indexes the layer, and 
consider the training 
sequence $x_1, x_2 , x_3, \_, x_5$
and the permutation order \texttt{3,1,5,2,4}.
In BERT, we compute attention scores to obtain
$\mathbf{h}_t^k$ from $\mathbf{h}_t^{k-1}$ 
for every $t$ (i.e., $t=1,\ldots,T$),
where $\mathbf{h}_4^{0}$ is the embedding for the mask symbol.
In XLNet, the attention scores for future words in the
permutation order are masked to 0.
%the attention computation
%is not performed on every pair of 
%tokens in the sequence (with
%masked tokens being replaced by a mask symbol), but
%only between a word and all other words 
%that appear before the word 
%for a given permutation order.
For example, when we compute $\mathbf{h}_1^k$,
only the attention score from $\mathbf{h}_3^{k-1}$ 
is considered (since the permutation order 
is \texttt{3,1,5,2,4}). For $\mathbf{h}_5^k$,
we use $\mathbf{h}_1^{k-1}$ and $\mathbf{h}_3^{k-1}$.
XLNet does not require a mask symbol embedding
since the attention score from a masked token
is always zeroed out 
with an attention mask (implemented as a matrix).
As a result, we can consider XLNet training
as masked language modeling
with stochastic attention masks in the encoder.

%Since the first layer of the query stream is initialized
%with a trainable parameter vector, this vector plays a role
%of the embedding for the mask symbol.
%Another difference is that 
%the permutation language modeling objective promotes 
%training on
%masked sequences with correlated prefixes from a sentence
%(e.g., predict $x_2$ from $x_1, \_ , x_3, \_, x_5$ 
%and predict $x_4$ from $x_1, x_2 , x_3, \_, x_5$).
%On the other hand, the masked 
%language modeling objective used in BERT
%samples sequences independently.
%In our example above, instead of having $x_1, \_ , x_3, \_, x_5$
%and $x_1, x_2 , x_3, \_, x_5$ as two training examples in a batch, they can
%be presented as one example, where we use representations 
%from the content stream of the previous target (e.g., the first masked sequence) 
%to compute the query stream for the next target (e.g., the second masked sequence).

%Since XLNet is trained on an objective similar 
%to the masked language modeling objective,
It is now straightforward to see that 
the permutation language modeling objective 
is an instance of Eq.\ref{eq:nce},
where $b$ is a target token $x_i$ and $a$ is a masked sequence 
$\hat{\boldsymbol{x}}_i = \{x_1,\ldots,\hat{x}_i,\ldots,x_T\}$.
Similar to BERT,
we have a Transformer encoder parameterized by
$\bm{\omega}$ and denote 
$g_{\bm{\omega}}(\boldsymbol{\hat{x}}_i) \in \mathbb{R}^d$
as a function that returns
the final hidden state 
corresponding to the $i$-th token 
(i.e., the masked token) after running $\boldsymbol{\hat{x}}_i$ through
the Transformer.
Let $g_{\bm{\psi}}: \mathcal{V} \rightarrow \mathbb{R}^d$ be 
a lookup function that 
maps each word type
into a vector and $\mathcal{\tilde{B}} = \mathcal{B}$ 
be the full vocabulary set $\mathcal{V}$.
The main difference between BERT and XLNet
is that the encoder that forms $g_{\bm{\omega}}$ used in XLNet
implements attention masking
based on a sampled permutation order when
building its representations.
In addition, XLNet and BERT also differ 
in the choice of $p(a,b)$
since each of them has its own masking procedure.
However, we can see that both XLNet and BERT maximize the
same objective.
%where the distribution of target tokens $b$ follows their respective
%since the target tokens used for training 
%How we sample
%The model of $p(a,b)$ is also different than BERT,
%since masked tokens in sequences used for training 
%follow a permutation order, whereas they are masked independently in BERT.

%As discussed above, the main difference between the 
%masked language modeling objective used in BERT and XLNet is
%in the joint distribution used to sample different views $p(a,b)$. 
%Given a sentence, masked tokens in XLNet 
%follow a permutation order, whereas they are independently masked in BERT.
%Since this is a standard language modeling loss with multiple factorizations,
%it straightforward to show that it is an instance of Eq.\ref{eq:nce}.
%Let $b$ be a word to be predicted (i.e., $x_t^{\boldsymbol{z}}$)
%and $a$ be its context under the permutation order $\boldsymbol{z}$,
%denoted by $\boldsymbol{x}_{<t}^{\boldsymbol{z}}$.
%Let $f_{\bm{\theta}}(a,b)$ be 
%$g_{\bm{\psi}}(b)^{\top}g_{\bm{\omega}}(a)$ and
%$\mathcal{\tilde{B}} = \mathcal{V}$ be the full vocabulary set,
%where $g_{\bm{\omega}}(a)$ is a Transformer (XL) encoder with two-stream
%self-attention, $g_{\bm{\psi}}(b): \mathcal{V} \rightarrow \mathbb{R}^d$ is a word embedding lookup function.
%In this framework, sampling different factorization orders 
%at training time corresponds to sampling from $p(a,b)$.

\section{InfoWord}
\label{sec:infobert}

Our analysis on Skip-Gram, BERT, and XLNet shows that their objective
functions are different instances of InfoNCE in Eq.\ref{eq:nce},
although they are typically trained using 
the entire vocabulary set for $\mathcal{\tilde{B}}$ 
instead of negative sampling.
These methods differ in how they choose which views 
of a sentence they use as $a$ and $b$,
the data distribution $p(a,b)$, and the architecture
of the encoder for computing $g_{\bm{\omega}}$, which
we summarize in Table~\ref{tbl:infonce}.
Seen under this unifying framework,
we can observe that progress in the field has largely 
been driven by using
a more powerful encoder to represent $g_{\bm{\omega}}$.
While we only provide derivations for Skip-gram, BERT, and XLNet,
it is straightforward
to show that other language-modeling-based 
pretraining-objectives such
as those used in ELMo \citep{elmo} and GPT-2 \citep{gpt2} 
can be formulated 
under this framework.

Our framework also allows us to draw connections to
other mutual information maximization 
representation learning methods
that have been successful 
in other domains (e.g., computer vision, audio processing,
reinforcement learning).
In this section, we discuss an example
derive insights to design a simple 
self-supervised objective 
for learning better language representations.

\begin{table}[t]
\small
    \centering
    \caption{Summary of methods as instances of contrastive learning. See text for details.} \label{tbl:infonce}
    \begin{tabular}{l|lllll}
     \toprule
     \textbf{Objective} & $a$ & $b$ & $p(a,b)$ & $g_{\bm{\omega}}$ & $g_{\bm{\psi}}$ \\%& $\mathcal{\tilde{B}}$ \\
     \midrule
     Skip-gram & word & word & word and its context & lookup & lookup \\%& $\mathcal{V}$\\
     %\midrule
     %GPT-2 & context & next word & context and next word & Transformer & lookup & $\mathcal{V}$\\
     MLM & context & masked word & masked tokens probability & Transformer & lookup \\%& $\mathcal{V}$\\
     NSP & sentence & sentence & (non-)consecutive sentences & Transformer & lookup \\%& $\mathcal{\tilde{S}}_2$\\
     XLNet & context & masked word & factorization permutation & TXL++ & lookup \\%& $\mathcal{V}$\\
     %\midrule
     %CPC & context & future word & context and next phrase & Transformer & lookup &$\mathcal{V}$\\
     DIM & context & masked $n$-grams & sentence and its $n$-grams & Transformer & not used \\%&$\mathcal{\tilde{S}}$\\
     %AMDIM & masked sentence & complement of $a$ & a sentence with two complementary masks & Transformer & Transformer &$\mathcal{\tilde{P}}$\\
     \bottomrule
    \end{tabular}
\end{table}

%\paragraph{Deep InfoMax.}
Deep InfoMax (DIM; \citealp{deepinfomax}) is 
a mutual information maximization 
based representation learning 
method for images. %that was proposed
%independently of CPC around the same time.
DIM shows that maximizing
the mutual information between an image representation
and local regions of the image improves the quality
of the representation.
The complete objective function that DIM maximizes consists
of multiple terms. Here, we focus on a term in the objective
that maximizes the mutual information
between local features and global features.
We describe the main idea of this objective for 
learning representations from a
one-dimensional sequence, although it is originally proposed 
to learn from a two-dimensional object.

Given a sequence $\boldsymbol{x} = \{x_1, x_2, \ldots, x_T\}$,
we consider the 
``global'' representation of the 
sequence to be the hidden state of the first token (assumed to be
a special start of sentence symbol)
after contextually 
encoding the sequence $g_{\bm{\omega}}(\boldsymbol{x})$,\footnote{Alternatively, 
the global representation can be
the averaged representations of words in the 
sequence 
although we do not explore this in our experiments.}
and the local representations to be 
the encoded representations of each word
in the sequence $g_{\bm{\psi}}(x_t)$.
We can use the contrastive learning framework to 
design a task that maximizes 
the mutual information between 
this global representation vector 
and its corresponding ``local'' representations using local 
representations from 
other sequences $g_{\bm{\psi}}({\hat{x}}_t)$
as negative samples.
This is analogous to training the global representation vector
of a sentence to
\emph{choose} which words appear in the sentence and which words
are from other sentences.\footnote{
We can see that this self-supervised task is related to the next 
sentence prediction objective in BERT. However, instead of
learning a global representation (assumed to be 
the representation of the first token in BERT) to be predictive of 
whether two sentences are consecutive sentences, it learns its global 
representation to be predictive of words in the original sentence.}
However, if we feed 
the original sequence $\boldsymbol{x}$ to the encoder
and take the hidden state of the first token 
as the global representation, the task becomes trivial
since the global representation is built using all the words in the sequence.
We instead use a masked sequence 
$a \coloneqq \boldsymbol{\hat{x}}_t = \{x_1, \ldots, \hat{x}_t, \ldots, x_T\}$
and $b \coloneqq x_t$.

%As learning only requires us to be able to take negative samples,
%the framework allows us to consider other local representations.
%Importantly, existing methods based on language modeling
%objectives uses a softmax function
State-of-the-art methods based 
on language modeling objectives
consider all negative samples 
since the second view of the input data 
(i.e., the part denoted by $b$ in Eq.~\ref{eq:nce}) 
that are used is simple and it consists of 
only a target word---hence the
size of the negative set is still manageable.
A major benefit of the contrastive learning framework 
is that we only
need to be able to take negative samples for training.
Instead of individual words, we can use
$n$-grams as the local representations.\footnote{Local image 
patches used in DIM are analogous to $n$-grams in a sentence.}
Denote an $n$-gram by $\boldsymbol{x}_{i:j}$ and a masked
sequence masked at position $i$ to $j$ by $\boldsymbol{\hat{x}}_{i:j}$
We define $\mathcal{I}_{\text{DIM}}$ as:
%\vspace{-0.2cm}
\begin{align*}
\mathcal{I}_{\text{DIM}} = \mathbb{E}_{p(\boldsymbol{\hat{x}}_{i:j},\boldsymbol{x}_{i:j})}\left[g_{\bm{\omega}}(\boldsymbol{\hat{x}}_{i:j})^{\top}g_{\bm{\omega}}(\boldsymbol{x}_{i:j}) - \log \sum_{\boldsymbol{\tilde{x}_{i:j}} \in \mathcal{\tilde{S}}} \exp(g_{\bm{\omega}}(\boldsymbol{\hat{x}}_{i:j})^{\top}g_{\bm{\omega}}(\boldsymbol{\tilde{x}}_{i:j})) \right],
\end{align*}
%\vspace{-0.5cm}
where $\boldsymbol{\hat{x}}_{i:j}$ is a sentence masked at position
$i$ to $j$, $\boldsymbol{x}_{i:j}$ is an $n$-gram spanning from $i$ to $j$, and
$\boldsymbol{\tilde{x}}_{i:j}$ is an $n$-gram from a set $\mathcal{\tilde{S}}$
that consists of the positive sample $\boldsymbol{x}_{i:j}$ and negative $n$-grams
from other sentences in the corpus.
We use one Transformer to encode both views, so
we do not need $g_{\bm{\psi}}$ here.

Since the main goal of representation learning is
to train an encoder parameterized by $\bm{\omega}$,
it is possible to combine multiple self-supervised tasks
into an objective function in the contrastive learning framework.
Our model, which we denote \textsc{InfoWord}, combines the above
objective---which is designed to improve sentence and
span representations---with a masked language modeling objective $\mathcal{I}_{\text{MLM}}$ for learning word representations.
The only difference between our masked language modeling objective 
and the standard masked language modeling objective is that
we use negative sampling to construct $\mathcal{\tilde{V}}$ by sampling
from the unigram distribution. We have:
\vspace{-0.2cm}
\begin{align*}
\mathcal{I}_{\text{MLM}} = \mathbb{E}_{p(\boldsymbol{\hat{x}}_i,x_i)}\left[g_{\bm{\omega}}(\boldsymbol{\hat{x}}_{i})^{\top}g_{\bm{\psi}}(x_i) - \log \sum_{\tilde{x}_i \in \mathcal{\tilde{V}}} \exp(g_{\bm{\omega}}(\boldsymbol{\hat{x}}_{i})^{\top}g_{\bm{\psi}}(\tilde{x}_{i})) \right],
\end{align*}
%\vspace{-0.5cm}
where $\boldsymbol{\hat{x}}_i$ a sentence masked at position $i$ and
$x_i$ is the $i$-th token in the sentence.

Our overall objective function is a weighted combination of the two terms above:
\begin{align*}
\mathcal{I}_{\textsc{InfoWord}} = \lambda_{\text{MLM}} \mathcal{I}_{\text{MLM}} + %\lambda_{\text{NSP}} I_{\text{NSP}} + \lambda_{\text{CPC}} I_{\text{CPC}} + 
\lambda_{\text{DIM}} \mathcal{I}_{\text{DIM}},
%+ \lambda_{\text{R}} \mathcal{R},
\end{align*}
where $\lambda_{\text{MLM}}$ and $\lambda_{\text{DIM}}$ are hyperparameters that 
balance the contribution of each term.

\section{Experiments}
\label{sec:experiments}
In this section, we evaluate the effects of
training masked language modeling with 
negative sampling and 
adding $\mathcal{I}_{\text{DIM}}$ to the quality
of learned representations.
%and conduct an ablation analysis
%to better understand the contribution of each component 
%in the InfoWord objective function in this section.

\subsection{Setup}
We largely follow the same experimental setup as the original BERT model \citep{bert}.
We have two Transformer architectures similar to BERT$_{\textsc{Base}}$
and BERT$_{\textsc{Large}}$.
BERT$_{\textsc{Base}}$ has
12 hidden layers, 768 hidden dimensions, and 12 
attention heads (110 million parameters);
whereas BERT$_{\textsc{Large}}$ has
24 hidden layers, 1024 hidden dimensions, and 16 
attention heads (340 million parameters).

For each of the Transformer variant above, we compare three models in our experiments:
\begin{itemize}
\item BERT: The original BERT model publicly available in \url{https://github.com/google-research/bert}.
\item BERT-NCE: Our reimplementation of BERT.
It differs from the original implementation
in several ways: (1) we only use the masked language modeling objective and
remove next sentence prediction, (2) we use negative sampling instead of softmax, and (3) we only use one sentence for each training example in a batch.
\item \textsc{InfoWord}: Our model described in \S\ref{sec:infobert}. The main difference between \textsc{InfoWord} and BERT-NCE is the addition of $\mathcal{I}_{\text{DIM}}$ to the objective function. We discuss how we mask the data for $\mathcal{I}_{\text{DIM}}$ in \S{\ref{sec:pretraining}}.
\end{itemize}

\subsection{Pretraining}
\label{sec:pretraining}
We use the same training corpora and 
apply the same preprocessing 
and tokenization as BERT.
%We create training examples using 
%a new masking procedure.
We create masked sequences for training 
with $\mathcal{I}_{\text{DIM}}$ as follows.
We iteratively sample $n$-grams from a sequence 
until the masking budget (15\% of the sequence length) 
has been spent. At each sampling iteration, 
we first sample the length of the 
$n$-gram (i.e., $n$ in $n$-grams) from a 
Gaussian distribution $\mathcal{N}(5,1)$ 
clipped at 1 (minimum length) and 10 (maximum length).
Since BERT tokenizes words into subwords,
we measure the $n$-gram
length at the word level 
and compute the masking budget at the subword level.
This procedure is inspired by the masking 
approach in \citet{spanbert}.
%\dycomment{lpk add masking rule here}

For negative sampling, we use words and $n$-grams 
from other sequences 
in the same batch as negative samples (for MLM and DIM
respectively).
There are approximately 70,000 subwords
and 10,000 $n$-grams (words and phrases) in a batch.
We discuss hyperparameter details in Appendix~\ref{appendix:hyperparams}.

%For $\text{LARGE}$ models, 
%we set $\lambda_{\text{MLM}}$ 
%to $1.0$ and $\lambda_{\text{DIM}}$ to $0.8$.
%We set $\lambda_{\text{MLM}} = \lambda_{\text{DIM}} = \lambda_{\text{AMDIM}} = 1$.
%--- scheduling ---

\subsection{Fine-tuning}
We evaluate
on two benchmarks: GLUE \citep{glue} and SQuAD\citep{squad}.
We train a task-specific decoder and fine-tune pretrained models
for each dataset that we consider.
We describe hyperparameter details in Appendix~\ref{appendix:hyperparams}.

\textbf{GLUE} is a set of natural 
language understanding tasks 
that includes sentiment analysis, 
linguistic acceptability, paraphrasing, 
and natural language inference. Each task is formulated as a
classification task. The tasks in GLUE are 
either a single-sentence
classification task or a sentence pair classification task.
We follow the same setup as the original BERT model 
and add a start of sentence symbol (i.e., the \texttt{CLS} symbol)
to every example and use a separator symbol (i.e., the \texttt{SEP} symbol)
to separate two concatenated sentences (for sentence pair classification tasks).
We add a linear transformation and a softmax layer to
predict the correct label (class) from the representation of the first token
of the sequence.
%We remove the segment embedding from the original BERT system as our implementation trained on single sequence of text.
%We do single-task fine-tuning for all the tasks.

\textbf{SQuAD} is a reading comprehension dataset constructed from
Wikipedia articles. We report results on SQuAD 1.1.
Here, we also follow the same setup as the original BERT model and predict an 
answer span---the start and end indices of the correct answer 
in the context. 
We use a standard span predictor as the decoder, which we
describe in details in Appendix~\ref{sec:qadecoder}.

%\begin{table}[t]
%    \centering
%    \caption{Summary of results on the SQuAD 1.1 and SQuAD 2.0 %datasets. (dev, wait for test)} \label{tbl:squadresults}
%    \begin{tabular}{lcc}
%     \toprule
%      \multirow{2}{*}{\textbf{Model}} & \multicolumn{2}{c}{\textbf{\textsc{SQuAD 1.1}}} \\ %\multicolumn{2}{c}{\textbf{\textsc{SQuAD 2.0}}} \\
%      & \textbf{$F_1$} & \textbf{EM} \\
%      \midrule
%      BERT & 88.5 & 80.8\\
%      Our BERT & 89.6 & 82.9\\
%      InfoWord & 90.2 & 83.4\\
%      \bottomrule
%    \end{tabular}
%    \vspace{-0.2cm}
%\end{table}

\subsection{Results}
\begin{figure}[t]
    \centering
    \includegraphics[scale=0.25]{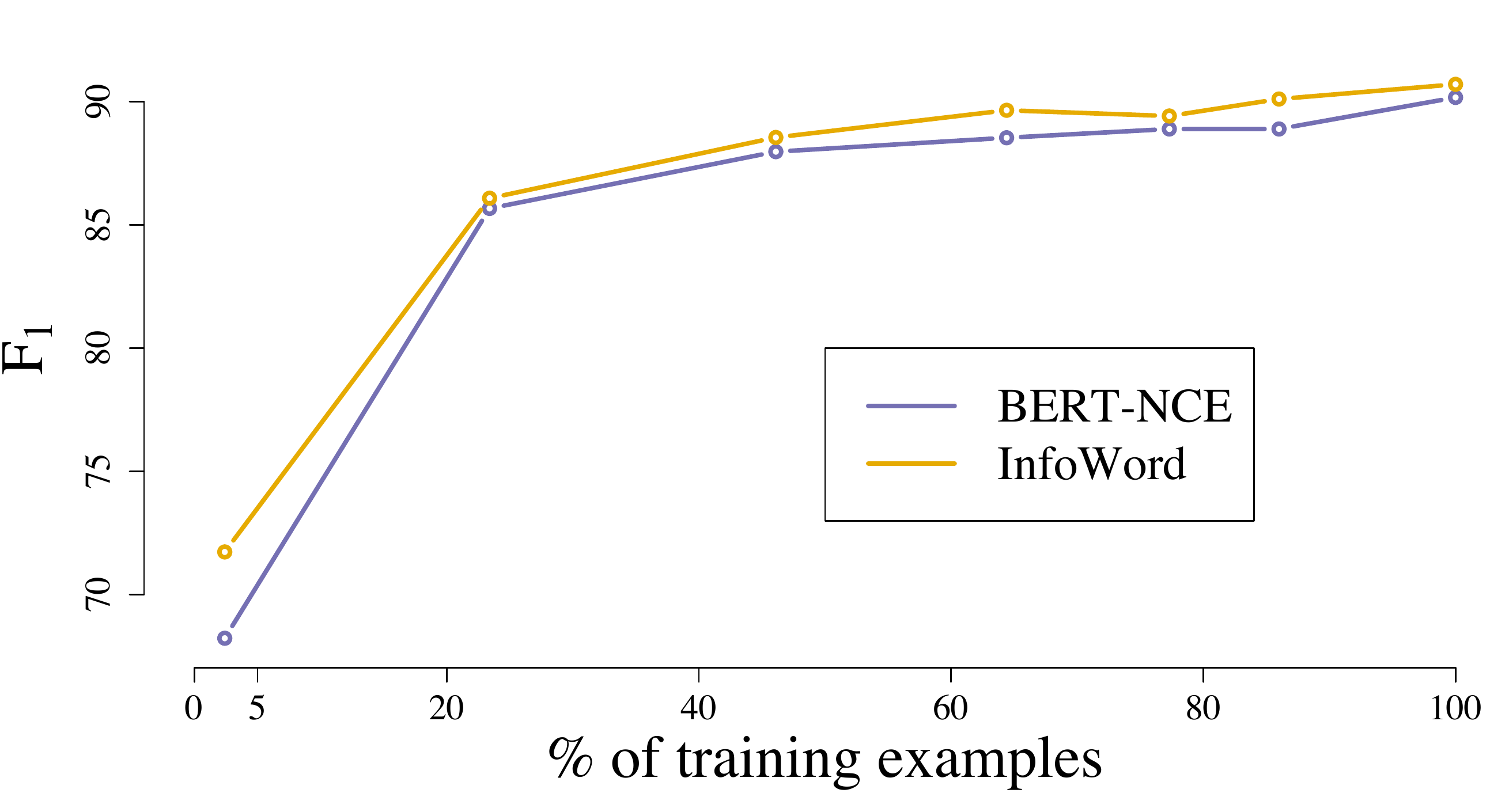}
    \includegraphics[scale=0.25]{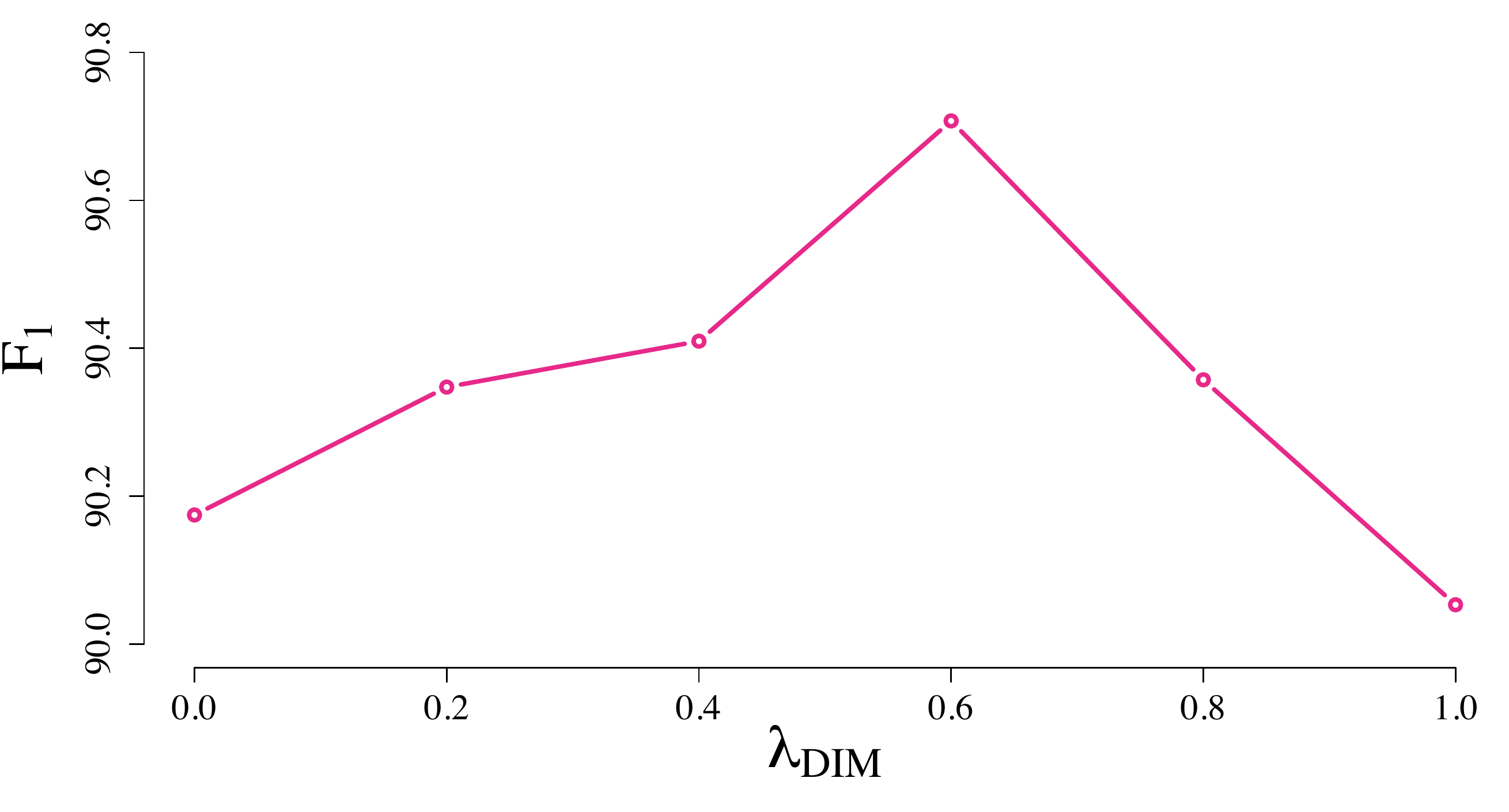}
\vspace{-0.3cm}
    \caption{The left plot shows $F_1$ scores
    of BERT-NCE and \textsc{InfoWord} as we increase the percentage
    of training examples on SQuAD (dev).
    The right plot shows $F_1$ scores 
    of \textsc{InfoWord} on SQuAD (dev) as a function of $\lambda_{\text{DIM}}$.}
    \label{fig:squad}
\vspace{-0.1cm}
\end{figure}

We show our main results in Table~\ref{tbl:results1} and Table~\ref{tbl:results2}.
Our BERT reimplementation with negative sampling 
underperforms the original BERT model on GLUE
but is significantly better on SQuAD. However, 
we think that the main reasons for this performance discrepancy are
the different masking procedures (we use span-based
masking instead of whole-word masking) and the different
ways training examples are presented to the model (we use
one consecutive sequence instead of two sequences 
separated by the separator symbol).
Comparing BERT-NCE and \textsc{InfoWord}, 
we observe the benefit
of the new self-supervised objective $\mathcal{I}_{\text{DIM}}$
(better overall GLUE and SQuAD results),
particularly on tasks such as
question answering and
linguistics acceptability
that seem to require understanding of longer phrases.
In order to better understand our model,
we investigate its performance with varying numbers of
training examples and different values of $\lambda_{\text{DIM}}$ 
on the SQuAD development set and show the results
in Figure~\ref{fig:squad} (for models with the \textsc{Base} configuration).
We can see that \textsc{InfoWord}
consistently outperforms BERT-NCE and 
the performance gap is biggest when the dataset is smallest, suggesting 
the benefit of having better pretrained representations 
when there are fewer training examples.

\begin{table}[t]
\setlength{\tabcolsep}{4.5pt}
\small
    \centering
    \caption{Summary of results on GLUE.} \label{tbl:results1}
    \begin{tabular}{ll|ccccccc|c}
     \toprule
      \multicolumn{2}{c|}{\multirow{2}{*}{\textbf{Model}}} & \multirow{2}{*}{\textbf{\textsc{CoLA}}} & \multirow{2}{*}{\textbf{\textsc{SST-2}}} & \multirow{2}{*}{\textbf{\textsc{MRPC}}} & \multirow{2}{*}{\textbf{\textsc{QQP}}} &
      \textbf{\textsc{MNLI}} & \multirow{2}{*}{\textbf{\textsc{QNLI}}} & \multirow{2}{*}{\textbf{\textsc{RTE}}} &
      \textbf{\textsc{GLUE}} \\
      &&&&&& \textbf{\textsc{(m/mm)}} & & & \textbf{\textsc{Avg}}\\
      \midrule
      \multirow{3}{*}{\rotatebox[origin=c]{90}{\textsc{Base}}} & BERT & 52.1 & 93.5 & 88.9 & 71.2 & 84.6/83.4 & 90.5 & 66.4 & 78.8  \\
      &BERT-NCE & 50.8 & 93.0 & 88.6  & 70.5 & 83.2/83.0 & 90.9 & 65.9 & 78.2 \\
      & \textsc{InfoWord} & 53.3 & 92.5 & 88.7 & 71.0 & 83.7/82.4 & 91.4 & 68.3 & \textbf{78.9} \\
      \midrule
      \multirow{3}{*}{\rotatebox[origin=c]{90}{\textsc{Large}}} &BERT &  60.5 & 94.9 & 89.3 & 72.1 & 86.7/85.9 & 92.7 & 70.1 & \textbf{81.5}\\
      &BERT-NCE & 54.7 & 93.1 & 89.5 & 71.2 & 85.8/85.0 & 92.7 & 72.5 & 80.6 \\
      & \textsc{InfoWord} & 57.5 & 94.2 & 90.2 & 71.3 & 85.8/84.8 & 92.6 & 72.0 & 81.1  \\
      \bottomrule
    \end{tabular}
    \vspace{-0.5cm}
\end{table}

\begin{table}[t]
\setlength{\tabcolsep}{4.5pt}
\small
    \centering
    \caption{Summary of results on SQuAD 1.1.} \label{tbl:results2}
    \begin{tabular}{ll|cc|cc}
     \toprule
      \multicolumn{2}{c|}{\multirow{2}{*}{\textbf{Model}}} & \multicolumn{2}{c|}{\textbf{\textsc{Dev}}} & \multicolumn{2}{c}{\textbf{\textsc{Test}}} \\
      && $F_1$ & EM & $F_1$ & EM \\
      \midrule
      \multirow{3}{*}{\rotatebox[origin=c]{90}{\textsc{Base}}} & BERT & 88.5 & 80.8 & - & -\\
      &BERT-NCE & 90.2 & 83.3 & 90.9 & 84.4 \\
      & \textsc{InfoWord} & \textbf{90.7} & \textbf{84.0} & \textbf{91.4} & \textbf{84.7}\\
      \midrule
      \multirow{3}{*}{\rotatebox[origin=c]{90}{\textsc{Large}}} &BERT & 90.9 & 84.1 & 91.3 & 84.3 \\
      &BERT-NCE & 92.0 & 85.9 & 92.7 & 86.6\\
      & \textsc{InfoWord} & \textbf{92.6} & \textbf{86.6} & \textbf{93.1} & \textbf{87.3}\\
      \bottomrule
    \end{tabular}
    \vspace{-0.5cm}
\end{table}

\subsection{Discussion}
\label{sec:discussion}

\paragraph{Span-based models.}
We show how to design a simple self-supervised task in the InfoNCE framework that improves downstream performance on
several datasets.
Learning language representations to predict 
contiguous masked tokens
has been explored in other context, and
the objective introduced in $\mathcal{I}_{\text{DIM}}$ is related to
these span-based models such as SpanBERT \citep{spanbert}
and MASS \citep{mass}. 
While our experimental goal is to demonstrate 
the benefit of contrastive learning for
constructing self-supervised tasks, we note that
\textsc{InfoWord} is simpler to train and exhibits similar trends
to SpanBERT that outperforms baseline models.
We leave exhaustive comparisons to these methods to future work.

\vspace{-0.2cm}
\paragraph{Mutual information maximization.}
A recent study has questioned whether the success of
InfoNCE as an objective function is due to its property
as a lower bound on mutual information and
provides an alternative hypothesis based on metric learning 
\citep{tschannen}. Regardless of the prevailing perspective,
InfoNCE is widely accepted as a good representation learning
objective, and formulating state-of-the-art language
representation learning methods under this framework offers
valuable insights that unifies many popular representation learning
methods.

\vspace{-0.2cm}
\paragraph{Regularization.}
Image representation learning methods often incorporate
a regularization term in its objective function 
to encourage
learned representations to look like a prior distribution
\citep{deepinfomax,amdim}.
This is useful for incorporating prior knowledge into a
representation learning model.
For example, 
%Given a distribution $p(\vect{x})$ that defines a prior 
%in a $d$-dimensional space 
%(e.g., a uniform distribution $[0,1]^d$)
%and a trainable vector $\bm{\phi} \in \mathbb{R}^d$,
the DeepInfoMax model has a term in its objective that encourages the 
learned representation from the
encoder to
match a uniform prior.
Regularization is not commonly used when learning language representations. 
Our analysis and the connection we draw to
representation learning methods used in other domains provide 
an insight into possible ways to incorporate
prior knowledge into language representation learning models.

\vspace{-0.2cm}
\paragraph{Future directions.}
The InfoNCE framework provides a holistic way to view progress
in language representation learning.
The framework is very flexible 
and suggests several directions that can be explored
to improve existing methods.
We show that progress in the field has been largely driven by
innovations in the encoder which forms $g_{\bm{\omega}}$.
InfoNCE is based on maximizing the mutual information between
different views of an input data, and it facilitates training
on structured views as long as we 
can perform negative sampling \citep{cpc,amdim}.
Our analysis demonstrates that existing methods based on language
modeling objectives only consider a single target word
as one of the views.
We think that incorporating more complex 
views (e.g., higher-order or skip $n$-grams, 
syntactic and semantic parses, etc.)
and designing appropriate self-supervised tasks is a
promising future direction. A related area that is also underexplored
is designing methods to obtain better negative samples.

%\begin{align*}
%\mathcal{R} = \arg\min_{\bm{\phi}} \arg\max_{\bm{\omega}} \mathbb{E}_{p(\vect{x})}\left[\bm{\phi}^{\top}\vect{x}\right] + \mathbb{E}_{\boldsymbol{x}}\left[\log(1 - \exp\left(\bm{\phi}^{\top}
%g_{\bm{\omega}}(\boldsymbol{x})\right)\right].
%\end{align*}

%\dycomment{greedy infomax}

%Augmented Multiscale DeepInfoMax (AMDIM; \citealp{amdim}) 
%extends DIM in
%several dimensions. One of the main contributions of AMDIM is to 
%introduce an objective to maximize mutual information 
%from multiple views of an image (e.g., two different
%patches of an image).

\section{Conclusion}
We analyzed
state-of-the-art language representation learning
methods from the perspective of mutual information maximization.
We provided a unifying view of classical
and modern word embedding models and showed 
how they relate to popular representation
learning methods used in other domains.
We used this framework
to construct a new self-supervised task
based on maximizing the mutual information
between the global representation and
local representations of a sentence.
We demonstrated
the benefit of this new task
via experiments on GLUE and SQuAD.

\bibliography{deepmind}
\bibliographystyle{iclr2020_conference}

\appendix
\section{Next Sentence Prediction}
\label{appendix:nsp}

We show that the next sentence prediction objective used in BERT 
is an instance of contrastive learning in this section.
In next sentence prediction, given two sentences $\boldsymbol{x}^1$
and $\boldsymbol{x}^2$, the task is to predict whether these are
two consecutive sentences or not. 
Training data for this task is created by
sampling a random second sentence $\boldsymbol{\hat{x}}^2$ 
from the corpus 
to be used as a negative example 50\% of the time.

Consider a discriminator (i.e., a classifier with parameters $\bm{\phi}$) that takes encoded representations of concatenated $\boldsymbol{x}^1$ 
and $\boldsymbol{x}^2$ and returns a score.
We denote this discriminator by $d_{\bm{\phi}}(\boldsymbol{x}^1,\boldsymbol{x}^2)$.
The next sentence prediction objective function is:
\begin{align*}
\mathbb{E}_{p(\boldsymbol{x}^1,\boldsymbol{x}^2)}\left[\log d_{\bm{\phi}}(g_{\bm{\omega}}([\boldsymbol{x}^1,\boldsymbol{x}^2)]) + \log (1 - d_{\bm{\phi}}(g_{\bm{\omega}}([\boldsymbol{x}^1,\boldsymbol{\tilde{x}}^2]))) \right].
\end{align*}
This objective function---which is used for training BERT---is 
known in the literature as
``local'' Noise Contrastive Estimation \citep{nce}.
Since summing over all possible negative sentences is intractable,
BERT approximates this by using a binary classifier to distinguish
real samples and noisy samples.

An alternative approximation to using a binary classifier
is to use ``global NCE'', which is what InfoNCE is based on. Here, we have:
\begin{align*}
\mathbb{E}_{p(\boldsymbol{x}^1,\boldsymbol{x}^2)}\left[\psi^{\top}g_{\bm{\omega}}([\boldsymbol{x}^1,\boldsymbol{x}^2)]) - \log \sum_{\tilde{x}^2 \in \mathcal{\tilde{X}}^2} \exp( \psi^{\top}(g_{\bm{\omega}}([\boldsymbol{x}^1,\boldsymbol{\tilde{x}}^2]))) \right],
\end{align*}
where we sample negative sentences from the corpus 
and combine it with the positive sentence
to construct $\mathcal{\tilde{X}}^2$. 
To make the connection
of this objective function with InfoNCE in Eq.~\ref{eq:nce} explicit, 
let $a$ and $b$ 
be two consecutive sentences
$\boldsymbol{x}_1$ and $\boldsymbol{x}_2$.
Let $f_{\bm{\theta}}(a,b)$ be $\bm{\psi}^{\top}g_{\bm{\omega}}([a,b])$, 
where $\bm{\psi} \in \mathbb{R}^d$ is a trainable parameter, 
$[a,b]$ denotes a concatenation of $a$ and $b$.
Consider a Transformer encoder parameterized by $\bm{\omega}$,
and let $g_{\bm{\omega}}([a,b]) \in \mathbb{R}^d$ be a
function that returns the final hidden state of the first token
after running the concatenated sequence to the Transformer.
%Let $\mathcal{\tilde{B}}$ be a set of size one of a random sentence
%sampled from the corpus $\boldsymbol{\hat{x}}_2$ that does not follow %$\boldsymbol{x}_1$, i.e., $\mathcal{\tilde{X}}_2$.
%Such a formulation of next sentence prediction maximizes Eq.~\ref{eq:nce}.
Note that the encoder that we want 
to learn only depends on $g_{\bm{\omega}}$, so both of these approximations
can be used for training next sentence prediction.

\section{Hyperparameters}
\label{appendix:hyperparams}
\paragraph{Pretraining.}
We use Adam \citep{adam} with $\beta_1=0.9$, $\beta_2=0.98$ and $\epsilon=1e-6$. The batch size for training is 1024 with a maximum sequence length of $512$. We train for 400,000 steps (including 18,000 warmup steps) with a weight decay rate of 0.01. 
We set the learning rate to $4e^{-4}$ for all variants of 
the $\textsc{Base}$ 
models and $1e^{-4}$ for the $\textsc{Large}$ models. 
We set $\lambda_{\text{MLM}}$ 
to $1.0$ and tune $\lambda_{\text{DIM}} \in \{0.4, 0.6, 0.8, 1.0\}$.

\paragraph{GLUE.} 
We set the 
maximum sequence length to 128.
For each GLUE task, 
we use the respective development set to choose 
the learning rate from $\{5e^{-6}, 1e^{-5}, 2e^{-5}, 3e^{-5}, 5e^{-5}\}$,
and the batch size from $\{16, 32\}$. 
The number of training epochs is set to 4 for CoLA 
and 10 for other tasks, following \citet{spanbert}.
We run each hyperparameter configuration 5 times
and evaluate the best model on the test set (once).

% the number of training epochs from $\{2, 3, 4\}$,

\paragraph{SQuAD.}
We set the maximum sequence length to 512
and train for 4 epochs.
%and use a sliding window of size $128$ 
%if the lengths are longer than $512$. 
We use the development set to choose the learning rate from 
$\{5e^{-6}, 1e^{-5}, 2e^{-5}, 3e^{-5}, 5e^{-5}\}$ and
the batch size from $\{16, 32\}$.

\section{Question Answering Decoder}
\label{sec:qadecoder}
We use a standard span predictor as follows.
Denote the length
of the context paragraph by $M$, 
and $\struct{x}^{\text{context}} = \{x^{\text{context}}_{1}, \ldots, x^{\text{context}}_{M}\}$.
Denote the encoded representation of the $m$-th token in the context
by $\vect{x}^{\text{context}}_{t,m}$.
The question answering decoder introduces 
two sets of parameters: $\vect{w}_{\text{start}}$
and $\vect{w}_{\text{end}}$.
The probability of each context token being the start of the answer
is computed as:
$p(\texttt{start} = x^{\text{context}}_{t,m} \mid \struct{x}_t) = \frac{\exp(\vect{w}_{\text{start}}^{\top}\vect{x}^{\text{context}}_{t,m})}{\sum_{n=0}^M \exp(\vect{w}_{\text{start}}^{\top}\vect{x}^{\text{context}}_{t,n})}$.
The probability of the end index of the answer is computed analogously using $\vect{w}_{\text{end}}$.
The predicted answer is the span with the highest probability
after multiplying the start and end probabilities. 

\end{document}

%% file: math_commands.tex
\usepackage{amsmath,amsfonts,bm}

\def\eqref#1{equation~\ref{#1}}
\def\1{\bm{1}}

\DeclareMathAlphabet{\mathsfit}{\encodingdefault}{\sfdefault}{m}{sl}
\SetMathAlphabet{\mathsfit}{bold}{\encodingdefault}{\sfdefault}{bx}{n}